\documentclass[11pt]{article}

\usepackage{jmlr2e}

\usepackage{booktabs}
\usepackage{amsmath}
\usepackage{amssymb}
\usepackage{amsmath}
\usepackage{comment}
\usepackage{lineno,hyperref}
\usepackage{verbatim}
\usepackage[ruled,vlined]{algorithm2e}

\usepackage{float}
\usepackage{xfrac}
\usepackage[all]{xy}
\usepackage{siunitx}
\usepackage{xcolor}
\usepackage{nth}
\usepackage{listings}

\usepackage{graphicx}
\usepackage{pgfplots}
\pgfplotsset{compat=1.14}

\usepackage{placeins}
\usepackage{longtable}
\usepackage{multirow}
\usepackage{booktabs}
\usepackage{centernot}

\pgfplotsset{compat=newest}
\pgfplotsset{plot coordinates/math parser=false}
\usepackage{tikzscale}
\usetikzlibrary{matrix,chains,positioning,decorations.pathreplacing,arrows}
\usepackage{tikz-qtree,tikz-qtree-compat}
\usetikzlibrary{calc}

\definecolor{BGgray}{gray}{0.9}
\definecolor{BGblue}{rgb}{0.941, 0.973, 1}
\newcommand{\ts}{\textsuperscript}

\ShortHeadings{The Need for Speed}{Malle, Giuliani, Kieseberg, Holzinger}
\firstpageno{1}

\begin{document}

\title{The Need for Speed of AI Applications\\ Performance Comparison of Native vs. Browser-based Algorithm Implementations}

\author{\name Bernd Malle\textsuperscript{1,2}  \email b.malle@hci-kdd.org\\
	\name Nicola Giuliani\textsuperscript{1}  \email n.giuliani@hci-kdd.org\\
	\name Peter Kieseberg\textsuperscript{1,2,3} \email pkieseberg@sba-research.org\\
	\name Andreas Holzinger\textsuperscript{1} \email a.holzinger@hci-kdd.org\\
	\\
	\addr \ts{1} Holzinger Group HCI-KDD, Institute for Medical Informatics,\\
   \noindent\hspace*{2mm} Statistics \& Documentation, Medical University Graz, Austria\\
	\addr \ts{2} SBA Research, Vienna, Austria\\
	\addr \ts{3} University of Applied Sciences St.P\"olten, Austria
}

\maketitle

\vspace{1cm}

\begin{abstract}
AI applications pose increasing demands on performance, so it is not surprising that the era of client-side distributed software is becoming important. On top of many AI applications already using mobile hardware, and even browsers for computationally demanding AI applications, we are already witnessing the emergence of client-side (federated) machine learning algorithms, driven by the interests of large corporations and startups alike. Apart from mathematical and algorithmic concerns, this trend especially demands new levels of computational efficiency from client environments. Consequently, this paper deals with the question of state-of-the-art performance by presenting a comparison study between native code and different browser-based implementations: \textit{JavaScript}, \textit{ASM.js} as well as \textit{WebAssembly} on a representative mix of algorithms. Our results show that current efforts in runtime optimization push the boundaries well towards (and even beyond) native binary performance. We analyze the results obtained and speculate on the reasons behind some surprises, rounding the paper off by outlining future possibilities as well as some of our own research efforts.

\vspace{0.5cm}

\begin{keywords}
Federated machine learning, Client side computing, distributed computing, In-browser computing, JavaScript Virtual Machine (JSVM), VM execution cycle, VM optimizations, parallelization, SIMD, WebWorkers, ASM.js, Web Assembly, GPGPU
\end{keywords}

\end{abstract}

\section{Introduction \& Motivation}
\label{sec:intro}
For some years now, distributed \& cloud computing as well as virtualized and container-based software have been buzzwords in the software engineering community. However, although these trends have helped making software infrastructure generally and AI applications specifically more stable, fault-tolerant, easy to control, scale, monitor and bill - one characteristic has never changed: the computationally most demanding parts of our software's logic, may it be recommenders, intelligent social network analysis or deep learning models, are still handled by enormous data-centers or GPU-clusters of incredible capacity \citep{ArmbrustEtAl:2010:CloudComputing}.

On the other hand, exponential hardware evolution has given us small mobile devices more capable than many servers ten years ago, enabling to run successful AI applications \citep{JordanMitchell:2015:MLtrendsScience}, which increasingly interact with the environment and with other smart devices \citep{MoritzEtAl:2017DistributedFramework} demanding even more performance \citep{StoicaEtAl:2017:ChallengesAI}. Following two questions arise:
\begin{enumerate}
  \item Could it be possible to create a \textit{new }internet, based not on proprietary data access and computation, but purely distributed computing models?
  \item Just like data-centers have shifted from specialized supercomputers to grids of commodity hardware, could the logical next step be the involvement of every laptop, tablet, phone and smart device on the planet into our future business models?
\end{enumerate}

The economic incentive is obvious: because access to powerful, centralized hardware will often be out of reach of a small organization, outsourcing the computational costs of an application to a network of clients is reasonable and would furthermore contribute to \textbf{vastly superior scalability} of the system, as any centralized server infrastructure could be reduced to its absolute minimum duties: 1) distributing the client-side code, as well as 2) routing, and 3) storing a certain amount of (global) data. Most of all such approaches are of greatest importance for the health domain, where \\
\\
In order to achieve such grand goals, we need:

\begin{itemize}
  \item new algorithms that can collaborate without relying too much on global (centralized) data;
  \item new communication protocols to make such collaboration efficient;
  \item new mathematical / machine learning models which are statistically robust under distributed uncertainty -- but most of all:
  \item raw power on the client side, not only in hardware but especially in software \& its underlying execution environments.
\end{itemize}

Ironically, one of the most progressive software execution and distribution environments to implement such a future solution is a technology straight from the 90's - the Web Browser. Code written for this universally available runtime carries several advantages: It is highly portable out-of-the-box, any update in the future is trivially easy to deploy, and the hurdle for entry is practically zero as Browsers come pre-installed on every consumer device and operating system.

In addition, the browser's \textit{sandbox model} (code running within has no access to data on the rest of the filesystem) guarantees better privacy then mobile Apps or traditional, installed software; distributed algorithms at the same time force developers to design new models which can efficiently learn on limited data. All of this will make it easier to comply with the ever-increasing demands of data protection regulations on all levels, exemplified by the new European General Data Protection Regulation (GDPR), the right to be forgotten \citep{MalleEtAl:2016:forgotten} and the need for trust \citep{Katharina:2018:Trust}.\\
\\
Finally -- due to modern platforms including Apache Cordova -- we are able to package web applications into practically any format, may it be iOS or Android Apps or desktop applications. Therefore, in this paper we concentrate on performance tests within the browser's Virtual Machine. Traditionally this implied JavaScript as the language of the (only) choice; in recent years however alternatives gained attraction - in many cases promising whole new levels of performance and efficiency.

\section{Selected Related Work}
\label{sec:related_work}

\cite{Fortuna2010} were pointing out that JS would be suitable as a target language for future compute-intensive applications, given the contemporary rapid evolution of JS virtual machines during the browser wars of the late 2000s. As JS offers no native concurrency, they implemented parallelism on a ''task'' level, where task is defined as at least one JS OpCode. They explored and contrasted loop-level parallelism to inter- and intra-function parallelism and measured speedups of 2.19 to 45 times (average), with significantly higher factors achieved on function parallelism. Another attempt on dynamic parallelization of JS Applications using a speculation mechanism was outlined in \citep{Mehrara2011}, where the authors used data flow analysis in addition to memory dependence analysis and memory access profiling. However, in real-world, web-based interaction-intensive applications, they were only able to measure a meek speedup of 2.19, in-line with the findings mentioned earlier.

Another attempt at automatic parallelization of JS code was made by \citep{WookKim2016}, who also point out the economic advantage of such an approach for startups and small development shops, as they are usually not well-equipped to write parallel code-bases themselves. Their implementation utilizing static compiler analysis detecting DOALL (loop-level) parallelism enabled multi-CPU execution by emitting JS LLVM IR (intermediate representation), resulting in a speedup of 3.02 on the Pixastic benchmarks as well as 1.92 times on real-world HTML5 benchmarks, requiring a compilation overhead of only 0.215\%.

Lower-level (non-thread) parallelism can be achieved via Single Instruction Multiple Data (SIMD), which is a technique to perform one CPU instruction on several pieces of data (streams) simultaneously. \cite{Jensen2015}, who also describe the spiral of increasing performance of JS (the Web platform in general) leading to ever more use cases, which in turn lead to higher performance requirements, implemented SIMD.js, introducing CPU-native vector datatypes such as float32$\times$4, float64$\times$2, and int32$\times$4 to the JavaScript language. They were able to achieve speedups between 2$\times$ and 8$\times$ with an average around 4$\times$ (unsurprisingly for 4-wide vectorization); however development on SIMD.js has since then been halted in favor of the upcoming WebAssembly's native SIMD capabilities.

Parallelizing JS applications can also be achieved by employing WebWorkers - light-weight execution units which can spread over multiple CPUs and do not share the same memory space, but communicate via message-passing. As modern browsers enable retrieving the number of logical cores but give no information about resource sharing / utilization, the authors of \citep{AdhiantoEtAl:2010:HPCtoolkit} suggested using a web worker pool to balance the amount of web workers dynamically depending on current resource utilization. In addition, thread-level speculation can extract parallel behavior from single-threaded JavaScript programs (akin to the methods already described). Their results show a predictable maximum speedup at 4 parallel Workers on a 4-core (8 virtual core) machine in the range of a 3$\times$-5$\times$ factor, with different browsers behaving within a $\sim$30\% margin.

A completely different approach and real breakthrough in speeding up web-based applications was achieved with the introduction of Emscripten \citep{Zakai2011}, which actually uses LLVM as its input format and outputs a statically typed subset of JS called ASM.js, enabling the JSVM to optimize code to a much higher degree than dynamic JS. In addition, it uses a flat memory layout like compiled binaries do, simulating pointer arithmetic etc., which makes compilation from C-like languages possible in the first place. Although criticized for opening up additional security holes by introducing potential risks like buffer-overflows which were already ''fixed'' by the JS object memory layout model, Emscripten has been extremely successful in speeding up JS to almost native speeds, while (in theory) enabling running any language in a browser by compiling its whole underlying runtime to ASM.js, as long as the runtime itself is written in a language compiling to LLVM.

\cite{Borins2014} used Emscripten to compile the Faust libraries C++ output to JS; Faust is a code generation engine which forms a functional approach to signal processing tasks; it is generally used to deploy a signal processor to various languages and platforms. In this work specifically it was employed to target the Web Audio API in order to create audio visualizations, add effects to audio etc. The whole pipeline encompassed code generation from Faust to C++ as well as compiling to ASM.js. Preliminary experimentation showed that the pipeline was functional, although exact time measurements were apparently not possible in 2014 since Emscripten was not able to translate C++ timing code to ASM properly (this is working as of 2018).


The third general approach of speeding up browser-based code lies in transferring it to the GPU (mostly via WebGL which in its current form is a port of OpenGL ES 2.1), thus making use of modern graphic hardware's great parallelization capabilities. This is usually done for image processing or video (games), but can also work in non-obvious cases - \citep{Ingole2015} report on experiments on dynamic graph data-structures, which is a better indicator of how GPU-parallelization will behave on general code-bases. They used \textit{Parallel.js} (and other libraries like Rivertrail~\citep{Barton2013} which are nowadays deprecated) to compute dynamic shortest paths on growing, shrinking and fully dynamic, directed, weighted graphs with positive edge weights. Their results show that for up to a few percent of edge addition / deletion, the dynamic version usually outperforms the static version, although the margins differ between graph structures and depend on the random choice of edges added / deleted.

As a general reflection on VMs, \cite{Wurthinger2014} point out that generic bytecode-VMs are usually slower than ones focused on a specific guest language, for their parsing- and optimization routines can be specifically tailored to the syntactic idiosyncrasies of one language (and patterns / idioms usually employed by its programmers). As a remedy they propose an architecture where guest language semantics are communicated to the host runtime via specific interpreters, then the host compiler optimizes the obtained intermediate code - only this host compiler as well as the garbage collector remain the same for all languages. They implemented this idea at Oracle Corporation in the form of \textit{Truffle} as guest runtime framework on top of the \textit{Graal} compiler running in the \textit{Hotspot} VM; at the time of their publication, they supported C, Java, Python, Javascript, R, Ruby and Smalltalk; performance measures had not yet been published.

Recently, the introduction of WebAssembly (WASM, \citep{Haas2017}) as a new low-level, bytecode-like standard including static types, a structured control flow as well as compact representation opens up new opportunities, as it is simply an abstraction over modern hardware, making it language-, hardware-, and platform-independent.

Moreover, it will be able to utilize CPU-based SIMD Vectorization as well as offer a thread-based model for concurrency on shared memory, which the JS-based WebWorker model was not able to deliver (it offered concurrency via message-passing only). As an entirely independent language, it can also be extended at any time without having to extend the underlying runtime, as is the case with ASM.js \& JS. 

Nevertheless, it can be included in JS programs easily and even compiled from LLVM-based bytecode via Emscripten, making it a breeze to experiment with by simply using existing (C/C++) code.

\section{Research goal}
\label{sec:main_hypothesis}

We posit that recent developments have made it feasible for browser-based solutions of non-trivial computational complexity to be competitive against native implementations.\\
\\
In order to show this, we selected a mix of algorithms which we implemented (or took implementations from Rosetta Code) in C++ as well as JavaScript and injected additional code for run-time measurement (wall time). We then compiled the C++ version via Emscripten \citep{Zakai:2011:Emscripten} to ASM.js as well as WebAssembly (WASM) and tested the results on native Linux (GCC-compiled binary), NodeJS (native JS, ASM, WASM) as well as Chrome, Firefox and Edge (ASM, WASM).
Our main interests in performing these experiments were:

\begin{itemize}
	\item to establish a baseline performance measure by executing binary (C++) code.
	\item to compare the performance of binary code to JS as well as Emscripten-compiled ASM.js / WASM.
	\item to establish an understanding of the different optimization possibilities for JS, ASM, WASM and develop insights as to which kind of computational tasks (numeric performance, memory-management, function-call efficiency, garbage-collection etc.) they would affect.
	\item to test those hypotheses on the performance measures we empirically observed and describe as well as speculate on any inexplicable deviation.
	\item to provide an outlook on future challenges and opportunities, predicated on our lessons learned.
\end{itemize}

\section{Algorithms}
\label{sect:algos_used}

In order to gain relevant insights into the performance of our chosen execution environments, we gathered a mixture of algorithms from very simple toy examples to a real-world graph problem employed in modern computer vision.

\subsection{Base tests}
\label{alg:base_tests}

Our toy examples consist of three test cases designed to elicit algorithmic performance w.r.t. 3 specific use-cases: 1) Basic memory management: Fill an array of length 1 million with random integers, 2) Function calls: Recursive Fibonacci of $n=40$ and 3) Numeric computations: Compare 10 million pairs of random integers.

\subsection{Floyd-Warshall}
\label{alg:floyd}

The Floyd-Warshall algorithm \citep{Floyd:1962:Algorithm} is an APSP (All-pair shorest-path) graph algorithm; given a graph $G(V,E)$ with $V$ being the set of vertices and $E$ being the set of edges, it works by choosing a vertex $k$ in the graph, then iterating over all possible pairs of vertices $(i,j)$, at each point deciding if there exists a shorter route between $(i,j)$ were they connected via $k$:

\begin{figure}[H]
\rule{\textwidth}{1pt}
\textbf{Algorithm: Floyd-Warshall APSP}
\begin{lstlisting}[mathescape=true]
function FWDense(graph) {
	i, j, k <- graph.vertices.length
  for (k = 0; k < V; k++) {
    for (i = 0; i < V; i++) {
      for (j = 0; j < V; j++) {
        if (graph[i][j] > graph[i][k] + graph[k][j]) {
          graph[i][j] = graph[i][k] + graph[k][j];
        }
      }
    }
  }
}
\end{lstlisting}
\vspace{-0.3cm}
\rule{\textwidth}{0.4pt}
\vspace{-0.8cm}
\caption{Floyd-Warshall APSP.}
\label{fig:huffman}
\end{figure}

We chose this algorithm due to it's simple implementation which relies on pure iterative performance, thus testing data structure as well as numerical efficiency.

\subsection{Huffman Coding}
\label{alg:huffman}

Huffman coding/encoding is a particular form of entropy encoding that is often used for lossless data compression \citep{HanEtAl:2015:HuffmanCoding}.
A prefix-free binary code with minimum expected codeword length is sought from a given set of symbols and their corresponding weights. The weights are usually proportional to probabilities.
Let \[L(C) = \sum_{i=1}^{n}w_i * length(c_i)\] be the weighted path length of code $C$. The goal consists of finding a code $C$ such that $L(C) \leq L(T)$ for all $T(A,W)$ where $C$ is defined as: \[ C(A,W) = (c_1,c_2,c_3,...,c_n),\] with the alphabet $A = \lbrace a_1,a_2,a_3,...,a_n \rbrace$ and the set of symbol weights $W = \lbrace w_1,w_2,w_3,...,w_n \rbrace$

\begin{figure}[H]
\rule{\textwidth}{1pt}
\textbf{Algorithm: Huffman(C)}
\begin{lstlisting}[mathescape=true]
n := |C|;
Q := C;
for i := 1 to n − 1 do
	allocate a new node z
	z.lef t := x := Extract-Min(Q);
	z.right := y := Extract-Min(Q);
	z.freq := x.freq + y.freq;
	Insert(Q, z);
end for
return Extract-Min(Q); {return the root of the tree}

\end{lstlisting}
\vspace{-0.3cm}
\rule{\textwidth}{0.4pt}
\vspace{-0.8cm}
\caption{Huffman(C) taken from \citep{Goemans2015}}
\label{fig:huffman}
\end{figure}

As Huffman's implementation is heavily dependent on its binary heap, whose operations depend on memory implementation but will be much faster than heavy copying / deletion of deep data structures, we suspected that JavaScript-based runtimes would fare relatively well in their own right.

\subsection{Permutation}
\label{alg:permutation}

The problem of permutation is simple to solve by dissecting a given input and combining its ''head'' with the inner permutations of its ''rest'', which can be done in a recursive fashion. For instance, given the input $abc$, the Function \textit{getPermutations} depicted below would first extract the letter $a$ and then combine it with all the permutations of the remaining string $bc$.

\begin{figure}[H]
\rule{\textwidth}{1pt}
\textbf{Algorithm: Permutation}
\begin{lstlisting}[mathescape=true]
function getPermutations(string text)
	define results as string[]
	if text is a single character
		add the character to results
		return results
	foreach char c in text
		define innerPermutations as string[]
		set innerPermutations to getPermutations (text without c)
		foreach string s in innerPermutations
			add c + s to results
	return results
\end{lstlisting}
\vspace{-0.3cm}
\rule{\textwidth}{0.4pt}
\vspace{-0.8cm}
\caption{String permutation}
\label{fig:string_perm}
\end{figure}

Since there are $n!$ possible permutations, this is also the expected runtime of the algorithm (without printing each permutation, which would result in $O(n*n!)$). As far as different language runtimes are concerned, we would suspect that those with efficient memory operations will have a significant advantage over others, since subset copying and joining will be the central operation.

\subsection{Fast Fourier Transform}
\label{alg:fft}

The Fast Fourier Transformation FFT Algorithm is a numerically very efficient algorithm to compute the Discrete Fourier Transformation (DFT) of a series of data samples (time series data). It goes back to Gauss (1805) and was quasi re-invented by Cooley \& Tukey (1965) \citep{CooleyTukey:1965:FFT}. The huge advantage is that the calculation of the coefficients of the DFT can be carried out iteratively and this reduces computational time dramatically \cite{CochranEtAl:1967:FFT}. A detailed description can be found in \cite{PuschelMoura:2008:FastAlgorithms}\\

Here a rapid explanation of the basics:

Let $f :\Bbb R \to \Bbb C$ a continuously function, where $R$ represents the set of real numbers
and $C$ the complex numbers. The {\it Fourier Transform} of
$f$ is given by
\begin{equation} \label{E1}
H(\nu) = \int_{-\infty}^{\infty}f(t)e^{- 2\pi i\nu t} dt,
\end{equation}
where $i$ is the imaginary unity and $\nu$ the frequency.\\

In most of practical situations, the function $f$ is given in discrete form as a
finite values collection $f(x_0), f(x_1), \dots ,f(x_{N-1})$ with $N \in \Bbb
N$, where  $\Bbb N$ is the set of natural numbers and
\{$x_0, x_1, \dots, x_{N-1}$\} is one partition on an real interval $[a, b]$ and $x_k = a+\frac{b-a}{N}k$, for
$0 \leq k \leq N-1$. In problems that imply numerical calculation, instead the equation (\ref{E1}) we use
the sum partial

\begin{equation} \label{E2}
H(k)={1\over N} \sum_{j=0}^{N-1} f(x_j)e^{- 2\pi ikj/N},
\end{equation}
designated  {\it Discrete Fourier Transform} (DFT) of $f$ over the interval $[a, b]$. If $f$ is a function
defined on the interval $[0,2\pi]$ of real value with period
$2\pi$,  the values $H(k)$, by
$k=0,\dots , N-1,$ can be interpreted as the coefficients $c_k$
\begin{equation} \label{coefFourier}
c_k=\frac{1}{N}\sum_{j=0}^{N-1}f(x_j)e^{-i\frac{2\pi k}{N}j}, \qquad \hbox{$0 \leq k \leq N-1$
}
\end{equation}
of a  {\it exponential polynomial}
\begin{equation} \label{E3}
p(x) = \sum_{k=0}^{N-1} c_k e^{ikx},
\end{equation}
where $x\in \Bbb R$,  $c_k \in \Bbb C,$ y $k=0,1,\dots, N-1,$  which interpolate
$f$ in the values
$f(x_0),f(x_1),\ldots,f(x_{N-1}).$\\

The Discrete Fourier Transform of $f$ over the  partition $\{x_0, x_1, \dots, x_{N-1}\},$
is defined as the operator $$DFT: \Bbb C^N\longrightarrow \Bbb C^N$$ such that
\begin{center}
	$[c_0, c_1,\ldots, c_{N-1}]^T=DFT\Big([f(x_0), f(x_1),\ldots,f(x_{N-1})]^T\Big)$
\end{center}

Technically, this is a divide \& conquer algorithm based on multi-branched recursion, i.e. it breaks down a problem into sub-problems until the problem becomes so simple that it can be solved directly.

\begin{figure}[H]
\rule{\textwidth}{1pt}
\textbf{Algorithm: Fast Fourier Transform, recursive}
\begin{lstlisting}[mathescape=true]
function y = fft_rec(x)
n = length(x);
if n == 1
	y = x;
else
	m = n/2;
	y_top = fft_rec(x(1:2:(n-1)));
	y_bottom = fft_rec(x(2:2:n));
	d = exp(-2 * pi * i / n) .^ (0:m-1);
	z = d .* y_bottom;
	y = [ y_top + z , y_top - z ];
end
\end{lstlisting}
\vspace{-0.3cm}
\rule{\textwidth}{0.4pt}
\vspace{-0.8cm}
\caption{FFT, taken from \citep{Stefan2008}}
\label{fig:fft_iterative}
\end{figure}

Since FFT is mostly about numerical computations we suspected that JS runtimes would handle this scenario relatively well, even compared to compiled code.

\subsection{Min Cut Max Flow}
\label{subsec:mcmf}

In order to obtain meaningful, relevant results from our experiments, we choose to add a non-trivial problem encompassing enough interesting 'moving parts' - memory management, computation intensity as well as a variety of (conditional) function call sequences - so that the resulting performance measures convey information applicable to real-world scenarios. We therefore decided on a theoretically simple graph problem (max-flow min-cut) which consists of only three computing stages, yet comprising enough internal complexity to make it interesting beyond toy samples.

\textbf{Theory}
Assuming a graph $G=\langle V,E \rangle$, where $V$ is a set of nodes and $E$ is the set of edges with positive edge weights connecting them, a cut is defined as the partition of the vertices of the graph $G$ into two disjoint sets. If the graph $G$ contains two distinct vertices $s$ and $t$, an $s-t-cut$ can be defined, which is a  partition into two disjoint sets $S$ and $T$ where $s$ is in $S$ and $t$ is in $T$. The cost of the $s-t-cut$ $C$ is the sum of weights of all edges that are connecting the two disjoint sets.
\[ c(S,T) = \sum_{\substack{e_{\lbrace i, j \rbrace} \in E, \\ |\lbrace i, j \rbrace \cap S | = 1 }} c_e \]\\
The minimum cut is then defined as the cut $c(S,T)$ amongst all possible cuts on $G$ with minimum cost.

If one considers a graph with two terminals $s,t$, one usually refers to these two as \emph{source} and \emph{sink} respectively. The max-flow problem asks how much \emph{flow} can be transferred from the \emph{source} to the \emph{sink}. This can be envisioned as the edges of the graphs being pipes, with their cost representing throughput capacity.

\cite{ford1962flows} stated that a maximum flow from $s$ to $t$ would saturate a set of edges in the graph, dividing it into two disjoint parts corresponding to a minimum cut. Thus, these two problems are equivalent, and the \emph{cost} of a min-cut is simultaneously the maximum flow.\\

\textbf{Graph Cuts in Computer Vision}

In computer vision a lot of problems can be formulated in terms of energy minimization - a labeling $f$ minimizing a given energy equation of the form $E(f) = \sum_{\lbrace p, q \rbrace \in N} V_{p,q}(f_p, f_q) + \sum_{p \in P} D_p(f_p)$ is sought. In the case of image segmentation the labels represent pixel intensities.

\textbf{$\alpha$ -Expansion Algorithm}

One algorithm for solving such energy equations is the $\alpha -Expansion$ algorithm by Boykov et al.~\citep{boykov2001em} An overview of the algorithm is given in figure~\ref{fig:EMAlgo}.  It is based on the computation of several minimum-cuts on specifically designed graphs $G_\alpha$, where each graph $G_\alpha$ corresponds to a labeling $f$. The cost of the cut on $G_{\alpha}$ corresponds to the energy $E(f)$. \\

The graph $G_{\alpha}$ changes for each label $\alpha$. Its set of vertices consists of a source ($\alpha$) and a sink ($\overline{\alpha}$) vertex, all image pixels $p$ in $P$ and all auxiliary vertices $a_{\lbrace p,q \rbrace}$. Auxiliary vertices are added for each pair of neighboring pixels $\lbrace p, q \rbrace \in N $ with $f_p \neq f_q$. \\


\begin{figure}[!ht]
\rule{\textwidth}{1pt}
\textbf{Algorithm: Boykov $\alpha$-expansion}
\begin{lstlisting}[mathescape=true]
set an arbitrary labeling $f$
set success := 0
for each label $\alpha \in \mathcal{L}$
	\text{} \quad 3.1 Find $\hat{f} = \text{arg min } E(\hat{f})$ among $\hat{f}$ within one $\alpha$-expansion move of $f$
	\text{} \quad 3.2 If $E(\hat{f}) < E(f)$, set $f = \hat{f} $ and success \text{ }:= 1
end for
if success == true goto 2
return $f$
\end{lstlisting}
\vspace{-0.3cm}
\rule{\textwidth}{0.4pt}
\vspace{-0.8cm}
\caption{$\alpha$-expansion algorithm \protect\citep{boykov2001em}}
\label{fig:EMAlgo}
\end{figure}

\textbf{Minimum-cut Algorithm}
The core of the previously presented $\alpha$ -expansion algorithm consists of computing minimum-cuts. Boykov et al. presented a suitable algorithm for solving the minimum-cut problem in~\citep{boykov2004mcmf}. It is based on finding augmenting paths and is comprised of three consecutive stages: \textbf{growth}, \textbf{augment} and \textbf{adoption}. During the computation four sets of vertices are maintained: the search trees $S$ and $T$, a set of active vertices $A$ and a set of orphan vertices $O$. Additionally the parent / child relations are kept.


Figure~\ref{fig:mincutMaxflowGrow} shows the detail of the growth stage. $TREE(p)$ denotes the tree of vertex $p$, $PARENT(q)$ means the parent of $p$.

After an augmenting path was found during the growth stage, the path gets augmented as described in figure~\ref{fig:mincutMaxflowAugment}. In the adoption stage new parents are sought for the orphan nodes.


\begin{figure}[!ht]
\rule{\textwidth}{1pt}
\textbf{Algorithm: Min-Cut growth stage}
\begin{lstlisting}[mathescape=true]
while $A \neq \emptyset$
	pick an active node $p \in A$
	for every neighbor $q$ such that $tree\_cap(p \rightarrow q) > 0$
		if $TREE(q) = \emptyset$ then add $q$ to search tree as an active node
		$TREE(q) := TREE(p), PARENT(q) := p, A := A \cup \lbrace q \rbrace$
		if $TREE(q) \neq \emptyset$ and $TREE(q) \neq TREE(p)$ return $P = PATH_{s \rightarrow t}$
	end for
	remove $p$ from $A$
end while
return $P = \emptyset$
\end{lstlisting}
\vspace{-0.3cm}
\rule{\textwidth}{0.4pt}
\vspace{-0.8cm}
\caption{Growth stage of the Min-Cut/Max-Flow algorithm \citep{boykov2004mcmf}}
\label{fig:mincutMaxflowGrow}
\end{figure}

\begin{figure}[!ht]
\rule{\textwidth}{1pt}
\textbf{Algorithm: Min-Cut augmentation stage}
\begin{lstlisting}[mathescape=true]
find the bottleneck capacity $\Delta$ on $P$
update the residual graph by pushing flow $\Delta$ through $P$
for each edge $(p, q)$ in $P$ that becomes saturated
	if $TREE(p) = TREE(q) = S$ then set $PARENT(q) := \emptyset$ and $O := P \cup \lbrace q \rbrace$
	if $TREE(p) = TREE(q) = T$ then set $PARENT(p) := \emptyset$ and $O := P \cup \lbrace p \rbrace$
end for
\end{lstlisting}
\vspace{-0.3cm}
\rule{\textwidth}{0.4pt}
\vspace{-0.8cm}
\caption{Augmentation stage of the Min-Cut / Max-Flow algorithm in~ \protect\cite{boykov2004mcmf}}
\label{fig:mincutMaxflowAugment}
\end{figure}

\begin{figure}[!ht]
\rule{\textwidth}{1pt}
\textbf{Algorithm: Min-Cut adoption stage}
\begin{lstlisting}[mathescape=true]
for all neighbors $q$ of $p$ such that $TREE(q) = TREE(p)$:
	if $tree\_cap(q \rightarrow p) > 0$ add $q$ to the active set $A$
	if $PARENT(q) = p$ add $q$ to the set of orphans $O$ and set $PARENT(q) := \emptyset$
$TREE(p) := \emptyset, A := A - \lbrace p \rbrace$
\end{lstlisting}
\vspace{-0.3cm}
\rule{\textwidth}{0.4pt}
\vspace{-0.8cm}
\caption{Adoption stage of the Min-Cut / Max-Flow algorithm in~ \protect\cite{boykov2004mcmf}}
\label{fig:mincutMaxflowAdopt}
\end{figure}

\textbf{Preprocessing}
We conducted our experiments on electron microscopy images from human skin. These images were first converted to gray-scale and thresholded to obtain a binary segmentation. Subsequently a graph was extracted from the binary image for label 255 as described above. Computing a minimum-cut on this graph would correspond to an $\alpha$-move for label 255. This graph was then saved and used for the following experiments.

\section{Experimental Setup}
\label{sect:experiments}

\subsection{Testing equipment}
\label{ssect:testing_equipment}
All tests were conducted on a 2017 Lenovo Carbon X1, 5th generation, with a Core i7-7500U, 16GB of LPDDR3 RAM and a PCIe-NVMe SSD. As C-compiler we utilized \textit{gcc (GCC) 7.2.1} with standard option flags ''-O3 -std=c++11''. For ASM.js / WASM we used \textit{Emscripten 1.37.32} with the same option flags ''-O3 --std=c++11'' (plus ''-s WASM=1'' for WASM). The string permutation sample had to be compiled with Emscripten option \textit{-s TOTAL\_MEMORY=268435456 -s ALLOW\_MEMORY\_GROWTH=1} in order to work at all, as simply using \textit{-s TOTAL\_MEMORY=XYZ} (no matter how great the amount) showed no effect in our experiments.

The Node ASM / WASM runs were conducted directly on the respective .asm.js / .wasm.js output files, where as the browser tests were conducted on the respective asm.html / wasm.html files under a standard Apache 2.24 local document root. The underlying operating system was \textit{Antergos (Arch) Linux with a 4.14.15-1-ARCH kernel}. For the Chrome runs, we used \textit{Chromium 64.0.3282.119 (Developer Build) (64-bit)} as provided by the Antergos (Arch) Linux distribution, for the Firefox benchmarks we used the Linux version \textit{58.0.1 (64-bit)}; finally, the Edge performance tests were conducted on the latest Edge Browser on \textit{Windows 10 Education, Build 1709}, on the same hardware as described above.

\subsection{Time measurement}
\label{ssect:time_measurement}

\begin{figure}[H]
	\begin{center}
		\vspace*{-0cm}
		\hspace*{-1cm}
		\includegraphics[width=1.1\textwidth,angle=0]{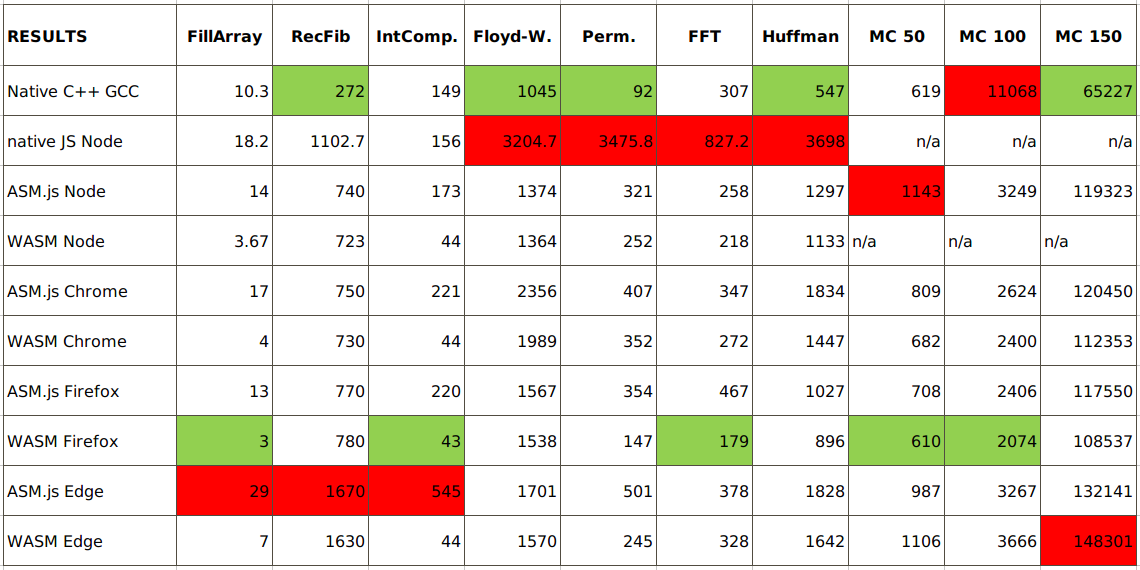}
		\caption{Overview table of detailed experimental results (runtime in milliseconds). Unsurprisingly, C++ compiled to binary wins in 50\% of experiments; the remaining 50\% go to Firefox's WASM implementation, which is comprehensible since the Mozilla foundation was first to introduce ASM.js / Emscripten and therefore has the most experience in runtime optimizations on the platform. It should be noted that WASM performance spread over different implementations was in no case dramatic (roughly within a factor of 2x), which is a testimony to the industry's good cooperation \& a positive sign for the future integrity of the platform.}
		\label{fig:results_table}
	\end{center}
\end{figure}

All timing was directly inserted into the code of our examples, to measure ''wall time'' from certain points of execution to others. For procedures that take too little time to conduct accurate measurements, we executed the procedure $n$ times - e.g. both the Fast Fourier Transformation as well as the Huffman Coding were done 100k times to arrive at a measurement. Each procedure was then repeated 10 times and the average taken to be the final measure as reported in Figure~\ref{fig:results_table}.

\section{Results and discussion}
\label{sec:results_discussion}

In order to understand the results better and give a starting point for their interpretation, we need to look at what happens when JS / ASM.js / WASM are executed (\citep{WASMFast}) in the JSVM (binary execution does not need further explanation):

\subsection{The interpretation / optimization / execution cycle:}
\label{ssect:op_cycle}

\begin{enumerate}
	\item \textbf{Fetching} means the downloading of the actual code representation. WASM has a slightly more compact form than JS, thus should load slightly faster.
	\item \textbf{Parsing / Validating} means ''reading'' and ''translating'' the code to an executable format, in our case either JS (incl. ASM.js) or WASM bytecode. WASM offers some advantages here, since it is already presented in a format much closer to the actual hardware-determined assembly (the virtual assembly format).
	\item \textbf{Monitoring} occurs during execution of JS/ASM.js (not WASM) in order to determine if an interpreted snippet of code (\textit{cold state}) is run frequently, in which case it is compiled via the base compiler into assembly (\textit{warn state}). For dynamically typed JS this state potentially produces a whole slew of different stumps (one for each combination of possible input value types) which then have to be chosen each time that snippet is called (depending on the ''current'' input value types). In case the stump is called sufficiently often, it enters the \textit{hot} phase.
	\item \textbf{Execution} is the main phase where code is actually run and produces effects. WASM is supposed to be slightly faster in this phase, since it's lower-level than (un-optimized) JS / ASM.js.
	\item \textbf{Optimization} For \textit{hot} snippets of code, the bytecode is further optimized making more or less aggressive assumptions (mostly about types again). This leads to less querying for each call (e.g. each iteration in a loop) but generates some overhead as optimizations need to be computed during runtime.
	\item \textbf{De-(Re)-optimizing / Bail-out}. In case assumptions do not hold - which can only be checked when a snippet of code is executed and some error occurs - the VM falls back to a baseline compilation or pure interpretation. This phase never occurs for WASM.
	\item \textbf{Garbage-collection}. Memory management; since WASM requires manual memory management, this phase is only necessary for JS/ASM.js. This phase can have a great effect on (absolute) runtime in case memory is heavily / carelessly freed. Especially the ''delete'' operator in JS is notoriously slow, but recursive algorithms operating on substrings may also be...
\end{enumerate}

\subsection{Results}
\label{ssect:results}

\subsection{Discussion}
\label{ssect:discussion}

We first note that our results for Recursive Fibonacci (of $n=40$), the Floyd-Warshall APSP Algorithm (on a graph of $|vertices| = 1k$) as well as 100k iterations of Huffman Coding (on a paragraph of lorem ipsum-based strings) yielded the expected results: Compiled C++ code came in fastest, with JS showing significantly worse performance (although easily within an order of magnitude). In all three cases, ASM.js performed significantly better than native JS, but worse than WASM which allows for higher optimization. We note that those algorithms are mostly computational and function-call oriented, meaning they do not heavily rely on memory management. In the only case where heap-operations are heavily involved (Huffman Coding) - we see that JS performance worsens comparatively - a sign that memory management in the VM is still slower than natively.

The results for String Permutations followed the expected patterns, but dramatically showed the impact of standard JS garbage collection on an exponential amount of substrings created and released: the performance dropped to about ~35x the time spent by the GCC compiled binary. ASM / WASM performed drastically better, the reason for which an be found in the flat-memory model of those implementations - chunks in the fixed-sized memory space are never really deallocated, but simply set inactive (undefined).

The first perplexing results lie in the run-times of filling an integer array with a million random numbers (FillArrayRand 1e6) as well as comparing ten million random integers pairs with one another (IntCompare 1e7). Here native (GCC-produced CPP) code is predictably beating ASM - although only by a slight margin when using the most recent version of NodeJS in ASM mode (factor ~1.2) - but blown away by WASM by up to a factor of ~3. This was so surprising that our first suggestion was Emscripten/Binaryen might be optimizing the whole procedure away; however subsequently added randomized access to the arrays (including output) alleviated this suspicion.
The same behavior, albeit to a lesser extent, can be observed in Fast Fourier Transformations (FFT 100k), where WASM manages to slightly beat the native binary. We note that all of those examples are heavily computation oriented, almost purely consisting of one giant number-crunching routine.

Therefore a possible interpretation for the WASM speedup could be SIMD Vectorization happening within the Emscripten pipeline; capabilities for SIMD are being actively developed for Chrome and Firefox, and would therefore also find their way into NodeJS. On the other hand, according to current sources \citep{SIMD} support for SIMD is not yet enabled in any standard build of any major JSVM. However, \citep{vectorizeJS2014} mention that their library \textit{Vectorize.js}, which can enhance normal JavaScript by applying 4-wide vectorization on CPU-heavy loops, is actually being used by Emscripten. While this does not guarantee that current ASM.js/WASM output is using vectorization, it is a strong hint that optimizations along those lines have been an effort for years. We will need to further investigate into possible intra-JSVM / WASM optimizations that might catapult performance beyond the CPP / binary baseline.

As far as our real-world scenario is concerned, we did not run the unoptimized JS version of this algorithm, since we found the sample code to be different enough between implementations as to make a fair comparison all but impossible. As for the GCC binary and (W)ASM, the two runs on a 50x50 / 150x150 pixel input image (graph) behaved as expected, although WASM on the latter was slightly outperformed by ASM. We can only assume that in this real-world code, optimization advantages regarding memory management (the graph was completely instantiated prior to the traversal) as well as static type assumptions did not play as great a role as expected. Furthermore, the task given most closely resembles the Floyd-Warshall (as in graph traversal) as well as RecFib40 (as in function-call centered) examples, in both of which ASM/WASM displayed similar behavior. All in all, we are satisfied to see results that 1) more or less matched our expectations and 2) behaved similarly to much smaller toy examples, since this indicates the technology is mature enough for real-world deployment.

The only case remaining inexplicable to us is that of the MinCut 100x100, in which GCC was outperformed by both ASM/WASM by a factor of more than 3x. A possible, yet unlikely, explanation could lie in a stark input abnormality in such a way that the resulting graph would allow traversal in a much different fashion from the other 2 samples: By accidentally presenting us with a graph in which the number of vertex discovery operations (involving a heap or other data-structure) are greatly reduced versus sheer numeric computations, we might get a result akin to the IntCompare one; however, the relatively good ASM.js performance speaks clearly against that; moreover, one could not attribute that behavior to implicit vectorization. In future work, we will strive to design input data to more complex algorithms in such a way as to uncover the explanatory factors hidden to us for now.

%

\begin{figure}[!ht]
	\begin{center}
		\vspace*{-0cm}
		\hspace*{-1cm}
		\includegraphics[width=1.1\textwidth,angle=0]{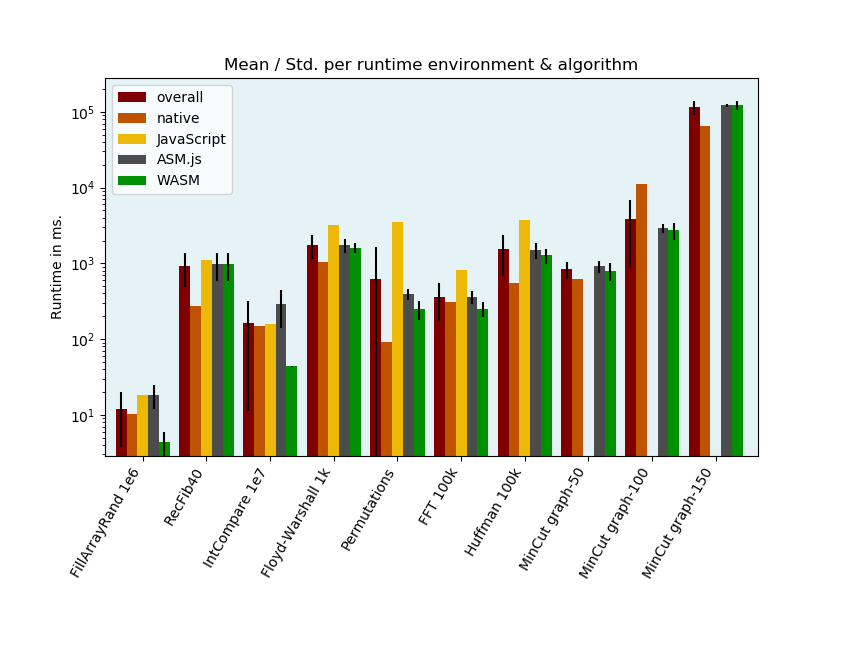}
		\caption{Overview plot of all results on one log scale including standard deviation. We note that overall performance deviation was most striking with FillArrayRand, IntCompare 1e7 as well as Permutations, but was less significant (although in no case negligible, as Figure~\ref{fig:results_per_group} clearly shows) in the case of Floyd Warshall 1k, FastFourierTransform, as well as MinCuts 50 \& 150}
		\label{fig:results_overall}
	\end{center}
\end{figure}

\begin{figure}[!t]
	\begin{center}
		\vspace*{-2cm}
		\hspace*{-1.2cm}
		\includegraphics[width=1.15\textwidth]{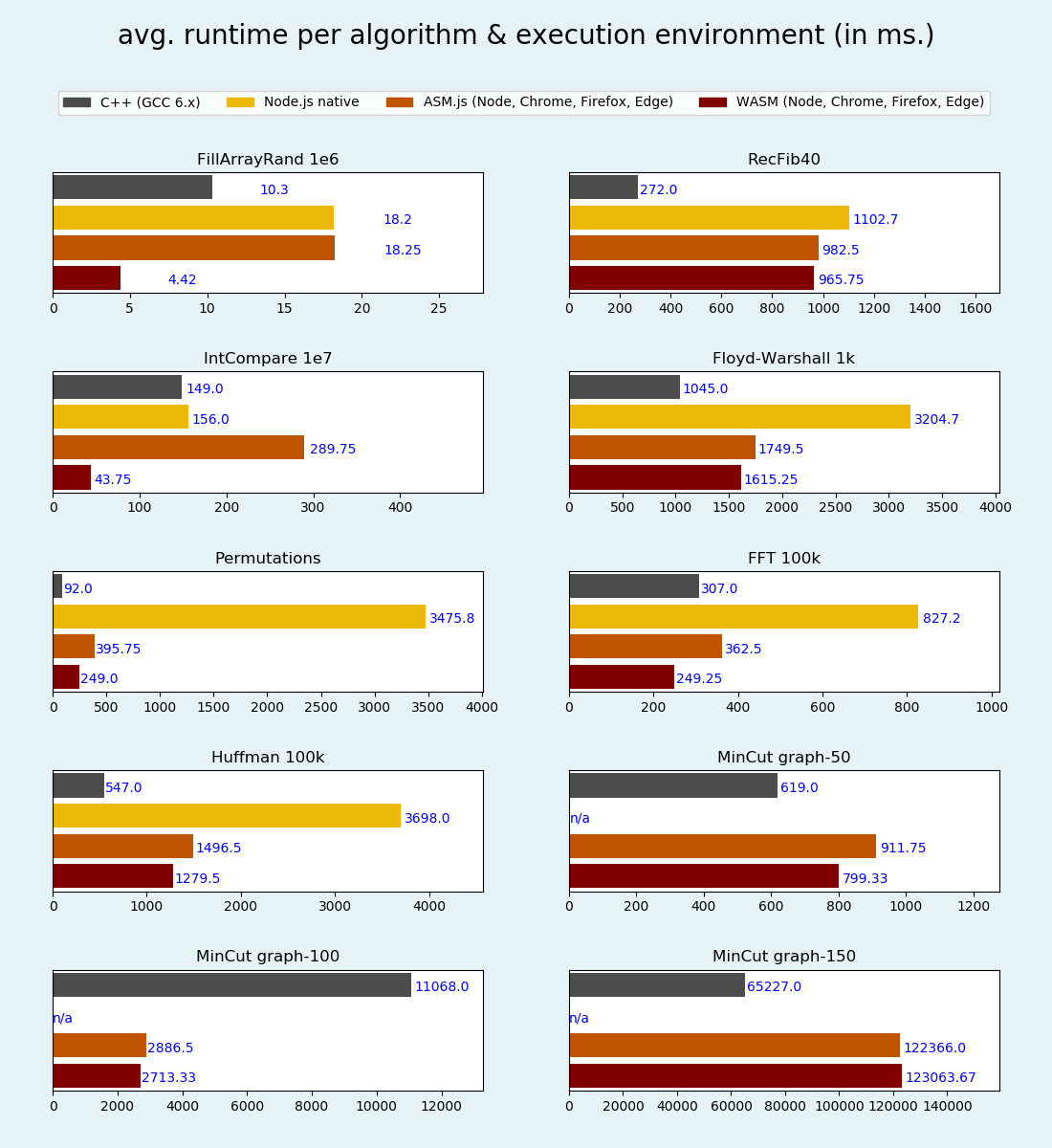}
		\caption{Performance of different runtime environments on the 10 chosen algorithms. These are the same results as in Figure~\ref{fig:results_overall}, but grouped per algorithm with each group on its own linear scale, thus relative runtime advantages can be depicted more intuitively.}
		\label{fig:results_per_group}
	\end{center}
\end{figure}

\section{Open problems \& Future challenges}
\label{sect:problems_challenges}

Although the results presented are very interesting and clearly point to areas of algorithmic programming in which a decisive speedup can be achieved employing ASM.js / WASM via Emscripten, not only are our current insights incomplete with respect to \emph{why} certain speedup factors can be achieved, nor have we exploited all possibilities to accelerate Browser-based software in general. Therefore, we will strive to improve and expand upon our research in four vital areas:

\begin{itemize}
	\item \textbf{Better profiling}. Deeper insights into the speedup of different parts of the execution pipeline for JS, ASM.js and WASM could help us target strengths as well as weaknesses of different platforms w.r.t. certain algorithmic requirements and programming styles a lot better; this would not only clear up the picture regarding explanatory factors but might also pave the way to deriving general rules as to what technology to utilize under certain conditions.
	\item \textbf{Future WASM features} like native SIMD vectorization and shared-memory concurrency can not been exploited as-of-date, at least not in a stable implementation. We will be closely monitoring the progress in the field and design experiments to optimally utilize any relevant upcoming features (e.g. parallel graph algorithms would be of great interest for client-side social network computations).
	\item \textbf{GPGPU}. Utilizing GPUs' massive parallelization capabilities even on modest hardware like integrated or mobile graphics could easily speed up parallelizable / vectorizable code beyond the speed limits of natively compiled C/C++. However, direct compilation from already (non-GPU enabled) code-bases is highly unlikely (except Emscripten's capacity for compiling OpenGL ES 2.1 code into WebGL, where the latter is a port of the former); therefore the additional effort of completely designing new code-bases for WebGL need to be taken into account. As a side-note, there exist libraries (like Tensorfire) who promise execution of Tensorflow models via WebGL - although this can not be regarded as 'speed-up' of traditional code, it would offer great potential for client-side classifiers once deep models have been obtained on more powerful hardware.
	\item \textbf{Federated Learning}. Since client-side / Browser-based computations are quickly entering the performance-range in which they are useful for even Machine Learning Tasks, their utilization in a ML-grid as proposed by Google's \textit{Federated Learning} initiative \citep{BonawitzEtAl:2016:FederatedGoogle} makes great sense even and especially for startups or mobile App developers, but also for decentralized health environments \citep{HolzingerEtAl:2017:AugmentedPathologist}, \citep{Shane:2017:VirtualAutopsyFirst}.  We are already working on a demonstration of this concept applied to distributed graphs and will gladly include our insights in a follow-up of this paper.
\end{itemize}

Overall, we are greatly enthusiastic about the current \& future possibilities
of client-side algorithmic computations and are looking forward to seeing any combination of the technologies at our disposal implemented in future applications (see e.g. \cite{MalleEtAl:2017:FederatedLearning}).

\section{Conclusion}
\label{sec:conclusion}

In this paper we presented a comparison study between the performance of native binary code, JavaScript, ASM.js as well as WASM on a selected set of algorithms we deemed representative for current challenges in client-side, distributed computing. Following a justification of this selection as well as some hypotheses on how different algorithms should perform in different execution environments, we presented experimental results and compared them to our prior expectations. We can conclude that although some scenarios played out exactly as we suspected, we were greatly surprised by others - especially WASM's capacity of (sometimes) outperforming native code by impressive margins will need further investigation into the internals of JSVM optimization. Above and beyond that, we point to the need of subsequent studies in parallelization as well as GPU computing to further expand on the work presented.


\bibliographystyle{unsrtnat}
\bibliography{Bibliography}

\end{document}